\documentclass[conference]{IEEEtran}
\IEEEoverridecommandlockouts
\usepackage{cite}
\usepackage{amsmath,amssymb,amsfonts}
\usepackage{algorithmic}
\usepackage{graphicx}
\usepackage{textcomp}
\usepackage{xcolor}
\def\BibTeX{{\rm B\kern-.05em{\sc i\kern-.025em b}\kern-.08em
    T\kern-.1667em\lower.7ex\hbox{E}\kern-.125emX}}

\usepackage{blindtext}
\usepackage{siunitx}
\usepackage{nicefrac}
\usepackage{graphics,graphicx}
\usepackage{amsmath}
\ifCLASSOPTIONcompsoc
\usepackage[caption=false,font=normalsize,labelfont=sf,textfont=sf]{subfig}
\else
  \usepackage[caption=false,font=footnotesize]{subfig}
\fi
\newcolumntype{M}[1]{>{\centering\arraybackslash}m{#1}}
\newcolumntype{R}[1]{>{\raggedleft\arraybackslash}m{#1}}
\newcolumntype{L}[1]{>{\raggedright\arraybackslash}m{#1}}
\def\midtilde@normaltilde{\texttildelow}
\DeclareMathOperator*{\minimize}{minimize}

\begin{document}
\bstctlcite{IEEEexample:BSTcontrol}
\title{FROST: Towards Energy-efficient AI-on-5G Platforms - A GPU Power Capping Evaluation}

\author{\IEEEauthorblockN{Ioannis Mavromatis\IEEEauthorrefmark{1},
Stefano De Feo\IEEEauthorrefmark{2}, Pietro Carnelli\IEEEauthorrefmark{1}, Robert J. Piechocki\IEEEauthorrefmark{2}, and 
Aftab Khan\IEEEauthorrefmark{1},\\
\IEEEauthorblockA{\IEEEauthorrefmark{1}Bristol Research \& Innovation Laboratory, Bristol, Toshiba Europe Ltd., UK\\ 
\IEEEauthorrefmark{2} Department of Electrical and Electronic Engineering, University of Bristol, UK\\}
Emails: \{ioannis.mavromatis,pietro.carnelli,aftab.khan\}@toshiba-bril.com}}


\maketitle

\begin{abstract}
The Open Radio Access Network (O-RAN) is a burgeoning market with projected growth in the upcoming years. RAN has the highest CAPEX impact on the network and, most importantly, consumes 73\% of its total energy. That makes it an ideal target for optimisation through the integration of Machine Learning (ML). However, the energy consumption of ML is frequently overlooked in such ecosystems. Our work addresses this critical aspect by presenting FROST -- Flexible Reconfiguration method with Online System Tuning -- a solution for energy-aware ML pipelines that adhere to O-RAN's specifications and principles. FROST is capable of profiling the energy consumption of an ML pipeline and optimising the hardware accordingly, thereby limiting the power draw. Our findings indicate that FROST can achieve energy savings of up to 26.4\% without compromising the model's accuracy or introducing significant time delays.
\end{abstract}

\begin{IEEEkeywords}
Machine Learning, Power Profiling, Power Capping, O-RAN, 5G, Sustainable AI, Green AI
\end{IEEEkeywords}

\section{Introduction}
5th-Generation (5G) network deployments are rapidly increasing, with over 1 billion connections globally as of 2022~\cite{website5G}. The Radio Access Network (RAN) of 5G is crucial in enabling a fully coordinated, multi-layer network infrastructure~\cite{5gRAN}. Open Radio Access Network (O-RAN) is an open specification for mobile fronthaul and midhaul networks built on cloud-native principles. It features disaggregated, virtualised, and software-based components connected via open standardised interfaces~\cite{oranOverview}. O-RAN ensures interoperability and standardisation, increased resilience and reconfigurability~\cite{oranResiliency}. Most importantly, it enables the integration of Machine Learning (ML)-enabled data-driven closed-loop controllers for optimising network and non-network related functions~\cite{5gMLapplications}. 

According to a report by GSMA Intelligence~\cite{gsmaIntelligence}, RAN consumes 73\% of the total energy of a 5G system. Compared to their 4G predecessors, 5G-RANs and O-RAN have significantly reduced energy (e.g., $39\%$ less energy is reported in~\cite{5gRANEnergyReduction}). By utilising ML and closed-loop control capabilities, O-RAN can further increase energy efficiency~\cite{understandingORAN}. Studies have shown that turning off underutilised basestations or deactivating antenna elements in a MIMO setup can improve energy efficiency by up to $22\%$~\cite{oranEnergyEfficiencyCellSwitchOff} and $18\%$~\cite{oranEnergyEfficiencyChannelSwitching}, respectively. Both studies conclude that it is vital to balance Quality-of-Service (QoS) and energy consumption, as there is always a trade-off between the two. 

In O-RAN literature, energy reductions are a priority, but the power consumed by the ML models is often overlooked. This is typical for most ML applications and services, with QoS enhancement frequently taking precedence over energy efficiency~\cite{energyPolicyAAAI}. Our work endeavoured to address this issue by developing FROST -- Flexible Reconfiguration method with Online System Tuning. FROST offers a practical solution for energy-conscious ML deployments in an O-RAN setting. It intelligently selects a suitable hardware configuration based on a given model, dataset, and training setup, considering the energy-delay cost benefits without compromising the model's accuracy. Our framework is readily accessible to the public via {\tt\small github.com/stefanodefeo/SustainableML}.

The energy consumption of ML is currently being extensively studied~\cite{greenAI}. Traditionally, energy savings have been estimated based on metrics such as the number of Multiply-Accumulate (MACs) operations, the number of ML model parameter weights stored in memory, or unique model features such as their layers~\cite{TYang2017-est,inaccurateMethods}. However, these methods are often inaccurate as they do not consider factors such as data collection and preprocessing, the hardware used, or introduced software optimisations~\cite{greenAI}. External power metering tools, both hardware- and software-based, offer a more accurate measurement of energy consumption~\cite{greenAI}. 

FROST is a software-based framework that runs in parallel to the ML pipeline, allowing for precise and comprehensive energy optimisations. As outlined in the O-RAN specification~\cite{oranMLFlow},  ensuring high accuracy, reliability, and timeliness is paramount in O-RAN ML pipelines. Thus, any energy reduction methods introduced must not compromise the model's accuracy or add significant overhead. 

While energy-aware ML techniques, such as quantisation and pruning, offer energy savings, they often come at the cost of the ML model's performance~\cite{greenAI}. Instead, FROST takes a different approach by introducing hardware optimisations that reduce the total energy consumed in an ML pipeline while ensuring unchanged model accuracy. FROST operates on the principle of power capping. Power capping constrains the power consumed by a CPU or a GPU, thus limiting energy consumption, albeit with a potential trade-off in hardware performance. While power capping has been widely deployed in software architectures for energy efficiency, its applicability to ML is still in the nascent stages, with only one existing work~\cite{gpucapping}. More importantly, it has never been proposed as a viable solution for an O-RAN system. This is the gap we are trying to bridge with FROST, providing a solution that can incorporated into O-RAN's specification with minimal effort. FROST, given a Convolutional Neural Network (CNN) model, a dataset, and a hardware setup, can find the optimal power capping configuration to minimise energy consumption.

The rest of the paper is organised as follows. Sec.~\ref{sec:powerBudget} provides a high-level overview of the ML interactions within O-RAN and discusses potential power budgeting strategies that can be incorporated. FROST is presented in Sec.~\ref{sec:methodology}, describing how the energy consumption can be profiled and how FROST can later decide on the optimum power capping configuration. This is followed by our results and discussion section (Sec.~\ref{sec:results} where we summarise our findings. Finally, Sec.~\ref{sec:conclusions} concludes this paper.

\section{Power Budgeting in O-RAN ML Pipelines}\label{sec:powerBudget}
O-RAN specification dictates the complete AI/ML lifecycle. ML's workflow is composed of six main steps: \textit{i}) data collection and processing, \textit{ii}) training; \textit{iii}) validation and publishing; \textit{iv}) deployment; \textit{v}) execution and inference, and \textit{vi}) continuous operation. This section describes more broadly O-RAN's ML operation and how power budgeting can be seamlessly integrated into existing O-RAN environments.

\subsection{RAN Architecture, O-RAN, and potential Use-Cases}
O-RAN is software-based and is built upon open and standardised interfaces, enabling easy reconfigurability and interoperability. It incorporates ML capabilities for on-the-fly system optimisations, reducing operational costs. Open Central Units (O-CUs) are split into the Control Plane and User Plane, accommodating multiple Open Distributed Units (O-DUs) and Open Radio Units (O-RUs), managing connection lifecycles, Service Data Adaptation Protocol (SDAP) and QoS. O-DUs control radio resources, Medium Access Control (MAC) and Radio Link Control (RLC), while O-RUs perform FFT and cyclic prefix operations. In addition, O-RAN specifies two RAN Intelligent Controller (RICs) that abstract the network, aggregate Key Performance Measurements (KPMs), and apply control policies via standardised interfaces (E1/E2, A1, O1/O2, etc.). Finally, the Service Management and Orchestration (SMO) platform introduces a data-driven closed-loop control, automating network- and non-network-related functions.

The non-Real-Time RIC (non-RT-RIC) specification outlines operations/optimisations for time scales larger than \SI{1}{\second}. The microservices running in non-RT-RIC (rAPPs) provide service, data management, and orchestration capabilities. These microservices expose services to data consumers and handle the AI/ML workflow, enhancing non-network functionality. The near-Real-Time RIC (near-RT-RIC) is deployed at the network edge and performs control loops with a periodicity between \SI{10}{\milli\second} and \SI{1}{\second}. The microservices within near-RT-RIC (xAPPs) govern the internal messaging, conflict mitigation, and subscription to functionalities via the exposed interfaces. Trained ML models can be deployed as xAPPs for orchestrating network-related functionality (e.g., network slicing). 

The O-RAN use-case whitepaper~\cite{oranWhitePaper} details various AI-based services and scenarios, such as context-based dynamic handover management for Vehicle-to-Everything (V2X), traffic steering, and flight path-based dynamic unmanned aerial vehicle (UAV) resource allocation etc. These services can be deployed within either of the two available RICs. As is evident from the research and O-RAN community, the potential for AI-enabled O-RAN services is far greater than that.

\begin{figure*}[t]
  \centering
  \includegraphics[width=0.83\textwidth]{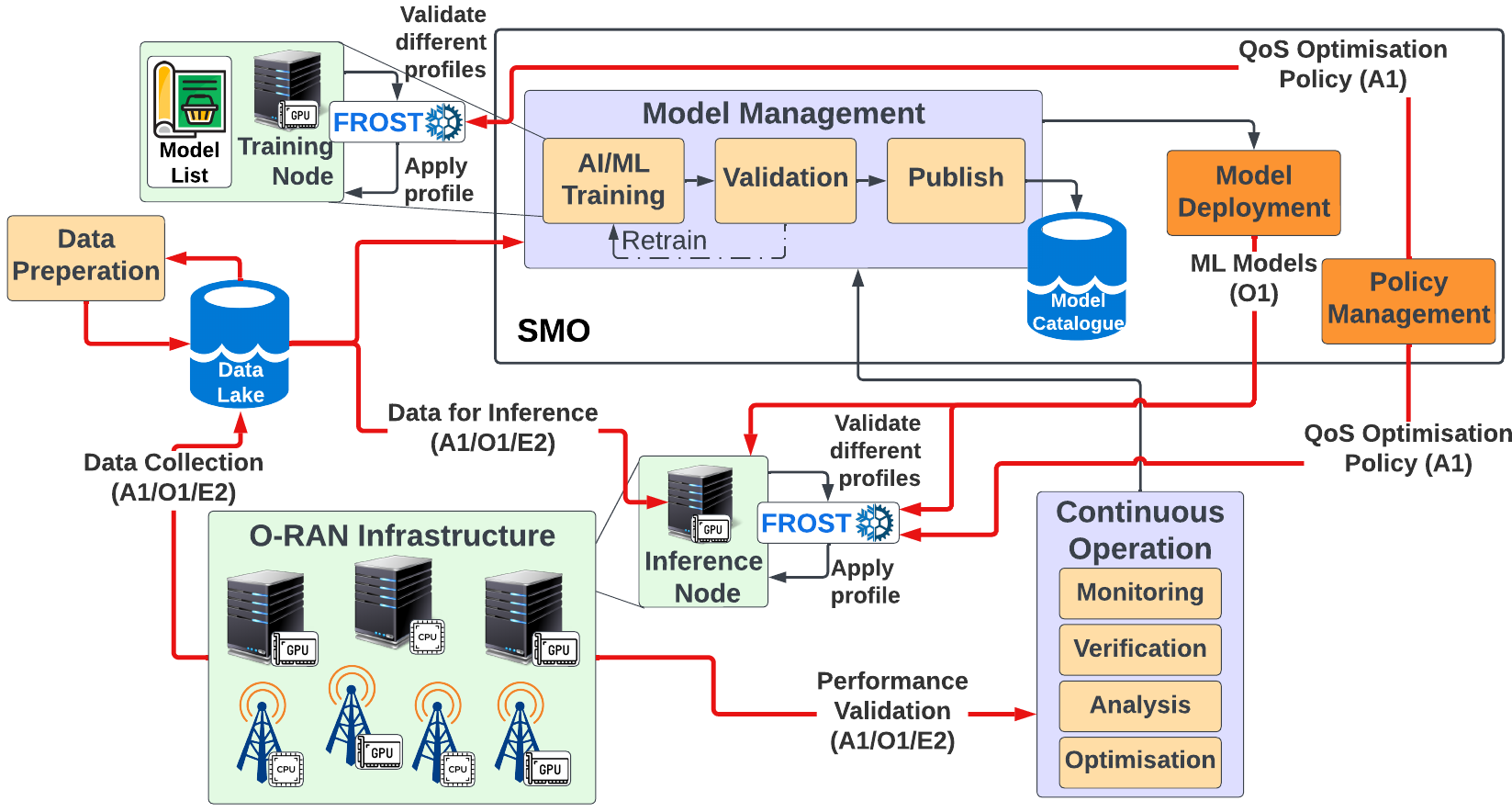}
  \caption{O-RAN ML Pipeline and the integration with FROST.}
  \label{fig:oranml}
\end{figure*}

\subsection{O-RAN and ML deployments}

Non-RT-RIC can manage and orchestrate the entire ML lifecycle. Data is collected via the O1, A1 and E2 interfaces and is stored in large data lakes or sent to inference hosts (O-RAN inference nodes)~\cite{oranMLFlow}. The data undergoes a preliminary pre-processing and preparation step before being used for offline training in O-RAN~\cite{oranMLFlow}. It is common to train multiple ML algorithms and identify the best one for accomplishing specific tasks~\cite{understandingORAN}. Once a model is trained, it is validated at the Non-RT-RIC, typically using a validation test dataset. This procedure either determines models needing retraining or ready to be published in an AI/ML catalogue. 


Models stored in the catalogue are subsequently deployed on the O-RAN inference hosts to perform online inference. These models run as containerised xAPPs or rAPPs. After deployment, they are continuously monitored and, if required, are fine-tuning online~\cite{oranMLFlow}. Fine-tuning is usually performed at the inference hosts (based on the hardware availability). Models could also be flagged for replacement/update by the SMO, and a new model is loaded from the catalogue.

O-RAN's main priority is to guarantee the accuracy and reliability of ML models.  However, the ML pipeline can be lengthy and involve a lot of trial-and-error. High-end GPUs often require more than \SI{300}{\watt} for operation. Complete server-grade systems can easily demand $\geq\SI{1000}{\watt}$. Unfortunately, the ML training pipeline, continuous inference, and online retraining can greatly increase power consumption in an O-RAN system and diminish the improvements shown in works like~\cite{oranEnergyEfficiencyCellSwitchOff,oranEnergyEfficiencyChannelSwitching}. FROST can provide a solution for the above, minimising the energy consumed by the ML lifecycle while ensuring high reliability and timely operation.

\subsection{Power Budgeting for O-RAN and FROST}
Power shifting is the dynamic setting of power budgets for individual system components to maintain a global power level. This is particularly important in an O-RAN deployment where multiple nodes may be involved in training or inference tasks, and optimising their power consumption locally or globally is necessary. 

Power capping and Dynamic Voltage and Frequency Scaling (DVFS) are two methods for achieving power shifting. Power capping reduces the voltage and frequency of a component, typically the CPU or GPU, when the maximum power threshold is exceeded. In contrast, DVFS dynamically adjusts the voltage and frequency to match the workload, providing more precise control than power capping and resulting in better energy savings.

However, implementing DVFS can be complex, as shown in~\cite{DVFS}. While frequency changes can reduce energy consumption, there is no direct correlation between frequency and energy consumption across GPU models, architectures, or manufacturers. Some GPUs have maximal energy efficiency at medium frequencies while others at higher operating ranges, making it challenging to develop hardware-agnostic optimisation strategies or make on-the-fly improvements without extensive trial-and-error to find an optimal configuration.

Furthermore, the available configuration parameters are affected by the deployment environment. For instance, authors in~\cite{amdWork} reduced the voltage levels of an AMD GPU, improving energy efficiency. However, this was only possible for AMD GPUs and not Nvidia, as Unix-based environments do not support the same functionality. On the contrary, power capping is readily available for all drivers and manufacturers, offering a more uniform solution across all domains.

An O-RAN system decouples the underlying hardware and host OS, operating cloud-natively. Furthermore, O-RAN deployments span multiple machines with varying characteristics, enabling interoperability and scalability. That makes power capping the only viable solution for O-RAN; FROST was created with the above in mind, enabling seamless integration across multiple platforms and devices. It can operate as a microservice deployed across all ML-related nodes, and taking as an input the energy-aware policies applied in the SMO, the model, and the data, it can reconfigure the hardware on-the-fly to minimise the power consumption. The reader can refer to Fig.~\ref{fig:oranml} for a high-level system diagram of our proposition. Nvidia is the de facto option for ML pipelines, providing better compatibility with ML software libraries such as TensorFlow or PyTorch and usually better performance due to the available CUDA cores. Our investigation will be based on a Nvidia-equipped setup; however, the proposed solution (modifying how power capping is implemented) is expected to work across hardware configurations from different manufacturers as well.

\section{Methodology}\label{sec:methodology}
FROST provides dual functionality. Firstly, it accurately measures real-time energy consumption with minimal overhead. Secondly, it dynamically adapts the system's configuration based on new policies received from the SMO.

\subsection{Software Energy Consumption Data}\label{subsec:energyAPIs}
Measuring energy consumption can be achieved through either hardware or software tools. While hardware solutions are more accurate~\cite{physicalMeter}, they are not practical for a deployment involving multiple interconnected devices, such as in O-RAN. Therefore, FROST is based on software-based alternatives, which can measure the energy consumption of various components, including the CPU, GPU, DRAM, disk I/Os, network I/Os, and peripheral devices. For our setup, we focus particularly on the  CPU, GPU, and DRAM, as there is no change in the hardware architecture, the network setup, or the data being exchanged.

Hardware manufacturers provide real-time power and energy consumption data through Model Specific Registers (MSRs), which can be accessed via well-defined APIs. For instance, the Nvidia management library (NVML) provides monitoring and management capabilities for Nvidia GPUs. Similarly, the Intel and Running Average Power Limit (RAPL) interface reports the power consumption of the CPU (all CPUs) and DRAM (only server-grade CPUs). Both interfaces have been validated and demonstrate a difference of $+/- \SI{5}{\watt}$ in absolute measurements, with similar trends in relative values~\cite{Nvidia2016,HDavid2010}. NVML reports raw measurements, while Intel's observed values are based on several MSR metrics. Our developed tool provides integration with both Intel's RAPL and Nvidia's NVML for measuring the power consumption of the CPU and the GPU of our test setups.

With regards to the DRAM's power draw, consumer CPUs do not provide MSR metrics. Therefore, for our investigation, we provide an estimation of DRAM's power consumption according to the number of DIMMs and the DIMM size. More specifically, according to~\cite{dramPowerConsumption}, each DIMM consumes $P_{\mathrm{DIMM}} = \nicefrac{1}{2} \ C V^2 f$, where $C$ is DRAM's capacitance, $V$ is the voltage applied, and $f$ is the operational frequency. While $V$ may change slightly during operation depending on the load, the change is almost unnoticeable on a macroscopic level. Additionally, newer generations of DRAMs typically operate at decreased voltages (e.g., going from DDR3 to DDR4), which decreases $C$ while $f$ increases. $C$ is directly related to each DRAM's cells (i.e., the DIMM size). For a DIMM, their size, frequency and model primarily dictate the power consumed, with the load not playing a significant factor. As a rule of thumb, we can use the following formula: $P_{\mathrm{DRAM}} = N_{\mathrm{DIMM}} \times \nicefrac{3}{8} \times S_{\mathrm{DIMM}}$, where $N_{\mathrm{DIMM}}$ is the number of DRAM DIMMs installed in the system, and $S_{\mathrm{DIMM}}$ is the size of each DIMM.

\subsection{Energy Measurements}\label{subsec:energyMeasurements}
In an ML pipeline, both training and inference processes consume a considerable amount of energy. To measure this energy consumption, we have defined two metrics, i.e., $E_{\mathrm{tr}}$, which is the total energy consumed during training, and $E_{\mathrm{in}}$, which is the total energy during inference. They are as follows:
\begin{equation}\label{eq:training}
    E_\mathrm{tr}=\int^{T_\mathrm{tr}}_{t=0} P_\mathrm{tr}(t) dt-\int^{T_m}_{t=0} P_\mathrm{idle}(t) dt
\end{equation}
\begin{equation}\label{eq:inference}
    E_\mathrm{in}=\int^{T_\mathrm{in}}_{t=0} P_\mathrm{in}(t) dt-\int^{T_m}_{t=0} P_\mathrm{idle}(t) dt
\end{equation}
where $T_\mathrm{tr}$ and $T_\mathrm{in}$ are the training and inference times, $T_{m}$ is a hardcoded time interval used for the idle experiment, and $P_\mathrm{tr}$, $P_\mathrm{in}$ and $P_{\mathrm{idle}}$ are the power measurements during training, testing and when the system is idle. For a given time $T$, the power $P$ is the sum of:
\begin{equation}
P(t) = \sum_{t=0}^{T} P_\mathrm{CPU}(t) + P_\mathrm{GPU}(t) + P_\mathrm{DRAM}(t)
\end{equation}
where $P_\mathrm{CPU}$, $P_\mathrm{GPU}$ and $P_\mathrm{DRAM}$ are the power consumption, taken in real-time for the CPU, GPU and DRAM, respectively.

\subsection{Power Profiler for an ML-enabled O-RAN Deployment}\label{subsec:profiler}
Based on the above measurements, we propose a power profiler that can be incorporated into an O-RAN deployment. Our solution can be utilised on all ML-enabled devices within the O-RAN network and has a dual purpose.  Its first objective is to measure the $P_\mathrm{CPU}$, $P_\mathrm{GPU}$ and $P_\mathrm{DRAM}$ (using the models APIs described in Sec.~\ref{subsec:energyAPIs}) without impeding the O-RAN network's operations or causing any extra load.  

As a second step, when a new ML model is introduced, the profiler tests various power limits for a brief period (\SI{30}{\second}) and chooses the most appropriate one. The default power limit of each device is set at $100\%$, which aligns with the device's Thermal Design Power (TDP). Lower limits than $100\%$ can be enforced through software. Though hardware boosts can force a device to operate momentarily over the limits, macroscopically, a lower number will decrease the energy consumed. Our profiler tests eight limits in the range of $30\% - 100\%$ at intervals of $10\%$ and determines the best strategy for the given model and setup.

The \SI{30}{\second} timeframe was chosen after calculating the correlation between the energy and training/inference times (which is linear - Sec.~\ref{sec:results}). For our hardware setups and the models tested, an epoch requires \textasciitilde\SIrange{7}{55}{\second} to run. A period of \SI{30}{\second} was considered adequate to allow enough samples/batches to be processed and provide a good indication of the energy consumed for a given model. The timing can be adjusted according to the hardware setup or the models used in a given use case.

\begin{figure*}[t]
  \centering
  \subfloat[Model accuracy VS energy.]{\includegraphics[width=0.28\linewidth]{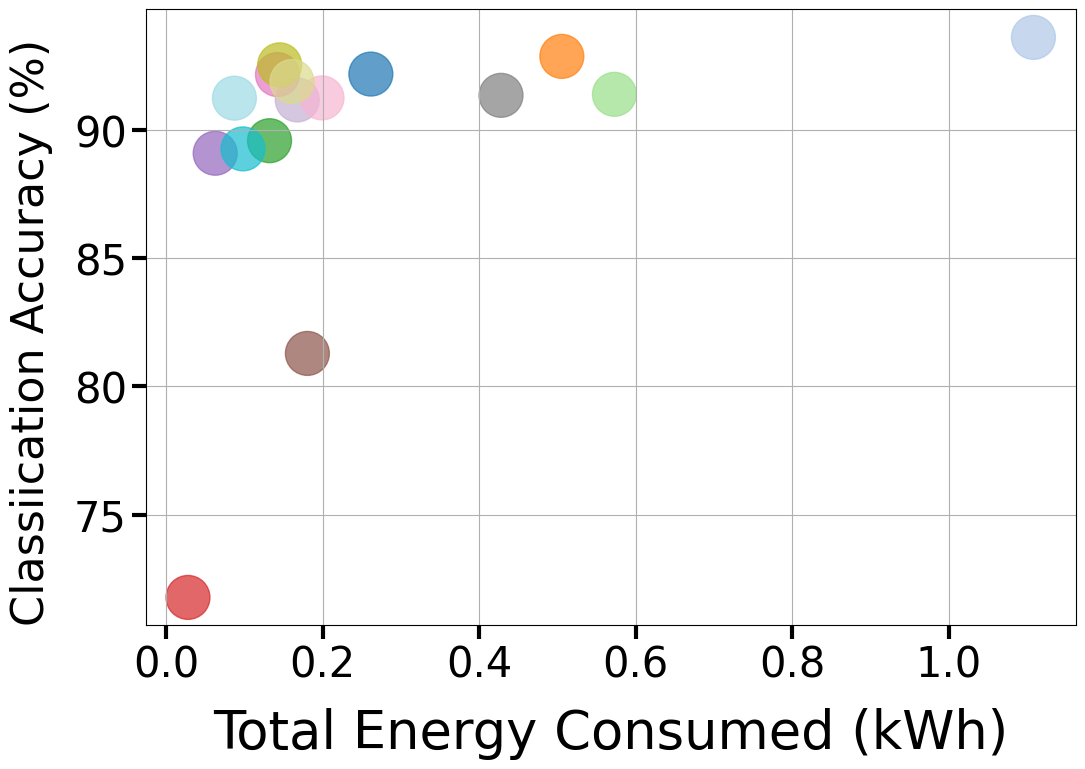}\label{subfig:acc_energy}}\hfill
  \subfloat[Training time VS energy.]{\includegraphics[width=0.28\linewidth]{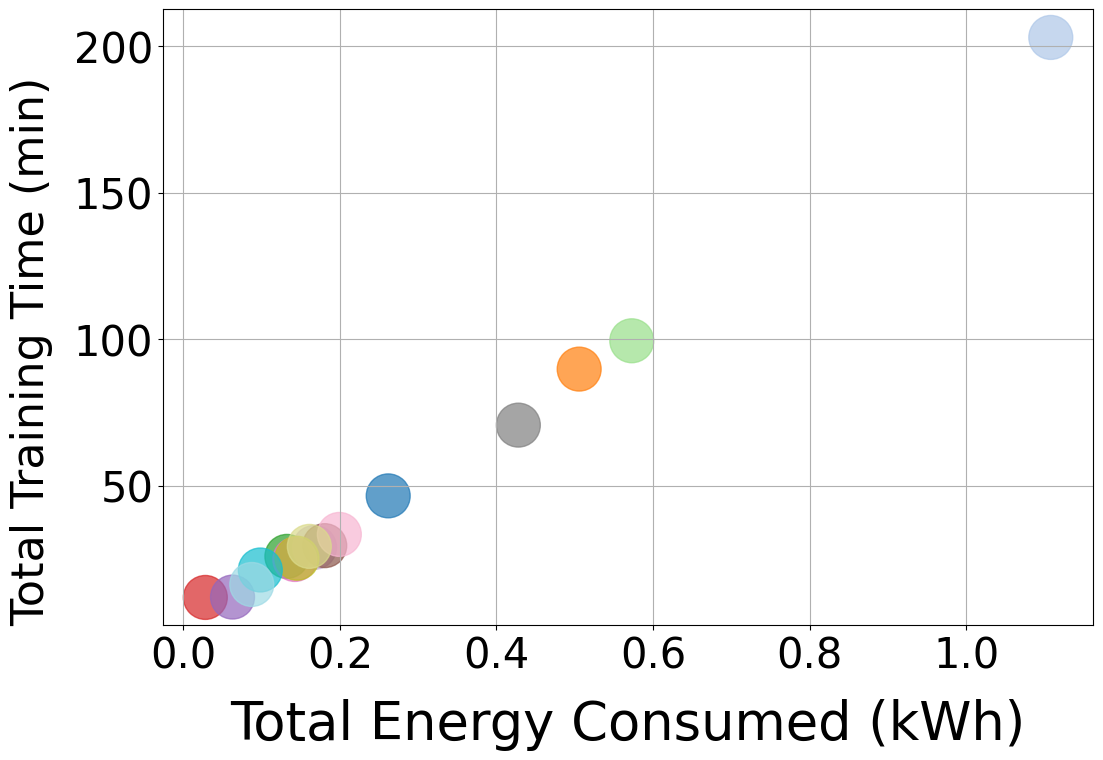}\label{subfig:train_energy}}\hfill
  \subfloat[GPU utilisation VS power draw.]{\includegraphics[width=0.28\linewidth]{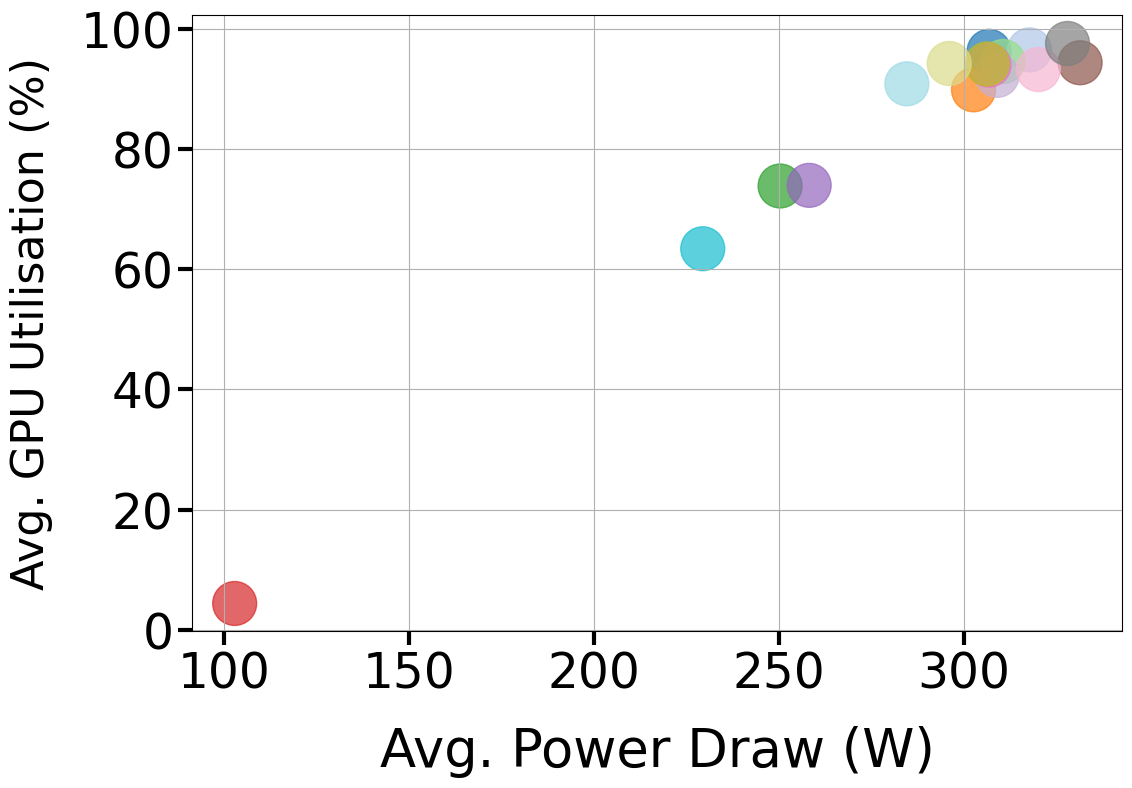}\label{subfig:util_power}}\hfill
  \subfloat{\includegraphics[width=0.1\linewidth]{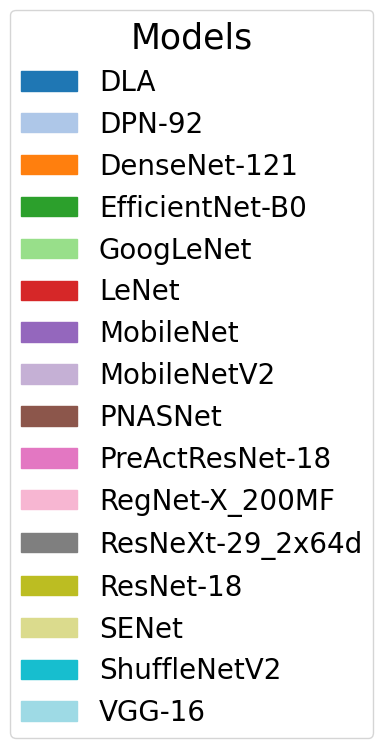}}
  \caption{Average statistics for all 16 models and 100 training epochs. GPU power draw and utilisation were averaged across 100 epochs.}
  \label{fig:init_investigation}
\end{figure*}

Based on the above-mentioned strategy, equations~\ref{eq:training} and ~\ref{eq:inference} become:
\begin{equation}
    E_\mathrm{tr}= 8\int^{T_\mathrm{pr}}_{t=0} P_\mathrm{pr}(t) dt + \int^{T_\mathrm{tr}}_{t=0} P_\mathrm{tr}(t) dt  -\int^{T_m}_{t=0} P_\mathrm{idle}(t) dt
\end{equation}
\begin{equation}
    E_\mathrm{in}= 8\int^{T_\mathrm{pr}}_{t=0} P_\mathrm{pr}(t) dt + \int^{T_\mathrm{in}}_{t=0} P_\mathrm{in}(t) dt  -\int^{T_m}_{t=0} P_\mathrm{idle}(t) dt
\end{equation}
where $P_\mathrm{pr}$ is the power consumed for each profiler test and runs for a given time $T_\mathrm{pr}$.

Our optimisation strategy is centred around the Energy-Delay Product (EDP)~\cite{edp}, where energy is the total energy consumption of the hardware and delay is the amount of time for executing applications. EDP is a commonly used metric for assessing the energy consumption of software applications~\cite{edp}. EDP is widely accepted, bridging software (algorithm design) and hardware, highlighting the trade-offs between energy and computational performance. Other metrics (e.g., carbon emissions, total cost of ownership, energy, time, etc.) are either too simplistic and do not account for various factors or are regionally dependent. EDP is calculated by multiplying the total energy and execution time. The goal of EDP minimisation is bi-objective optimisation. A fitting function $F$ is used to pick the best profile for any given configuration and model: 
\begin{align}
F(x) = ae^{bx-c} + d{\sigma}(ex-f) + g, \quad {\sigma}(x) = \frac{1}{1 + e^{-x}}
\end{align}
where $a,b,c,d,e,f,g$ are the coefficients that are fitted for each model, with $x$ being the values from each profile and ${\sigma}$ the logistic sigmoid function. This function was chosen after a fine-grained investigation of multiple statistical models and proved to be a good fit. The parameters were selected to enable effective shifting for both the exponential and logistic functions of the equation, thereby reducing the error. 

Having the eight initial profile values and the energy-per-sample, we fit $F(x)$ using the mean squared error. Minimising the error gives us a good fit for $F(x)$:
\begin{subequations}
\begin{alignat}{3}
\displaystyle{\minimize_{x\in N}} \quad e_{i}^2 \quad &= \quad && \frac{1}{N}\sum_{i=1}^{N} [y_i - F(x_i)]^2  \\
\text{subject to:} & && \ a,b,c,d,e,f,g
\end{alignat}
\end{subequations}
where $N$ is the total number of profiles tested. If the error drops below 5\%, we consider the line a good fit, giving us hyperparameters $a,b,c,d,e,f,g$. Later, using the downhill simplex algorithm, we find the minimum for $F(x)$, which determines the power limit that minimises the energy consumption for our specific setup.

Different applications may have varying QoS requirements regarding acceptable delays in a real-world O-RAN deployment. To accommodate that, we define a dependent parameter as $\mathrm{ED^mP}$, which allows us to use any combination of energy and execution time as a decision-making factor.  For instance, if we want to optimise for energy consumption, we can use a lower exponent (such as $\mathrm{ED^1P}$), but if we prioritise minimising delays, we can use a higher exponent (such as $\mathrm{ED^3P}$). These decisions can align with pre-defined QoS characteristics and be shaped as policies managed by the A1 Policy Management Service within the context of O-RAN.


\section{Results}\label{sec:results}
To evaluate our system, we conducted a thorough investigation to observe the behaviours of different ML pipelines using various existing deep-learning models. We picked: SimpleDLA, DPN (92), DenseNet (121), EfficientNet (B0), GoogLeNet, LeNet, MobileNet, MobileNetV2, PNASNet, PreActResNet (18), RegNet (X\_200MF), ResNet (18),  ResNeXt (29\_2x64d), SENet (18), ShuffleNetV2, and VGG (16), to capture a diverse range of architectures and sizes. Our experimentation consists of an initial energy investigation (Fig.~\ref{fig:init_investigation}) and an overhead evaluation. We also demonstrated the feasibility of our energy optimisation strategy and the tradeoffs it introduced. All experiments were conducted with the same set of hyperparameters. We used a batch size of $128$, a learning rate of $0.001$, ADAM optimiser and categorical cross entropy as our loss function.  We also fixed the seed to ensure consistency across different runs.

We used two different hardware configurations. The first included: Intel Core i7-8700K, \SI{64}{\giga\byte} DDR4 DRAM ($4 \times \SI{16}{\giga\byte}$ DIMMs at \SI{3600}{\mega\hertz}), Nvidia GeForce RTX 3080 (driver version: 525.105.17, CUDA version: 12.0). The second is: Intel Core i9-11900KF, \SI{128}{\giga\byte} DDR4 DRAM ($4 \times \SI{32}{\giga\byte}$ DIMMs at \SI{3200}{\mega\hertz}), Nvidia GeForce RTX 3090 (driver version: 530.30.02, CUDA version: 12.1). For the rest of the paper we refer to these configurations as setup no.1 and no.2 respectively. Due to limited space, we will present only a subset of our results and discuss the rest in the text.

\subsection{Initial Energy Performance Investigation}

We began our investigation by training all models for 100 epochs on the CIFAR-10 dataset and measuring the: accuracy, training time, energy consumption, and GPU utilisation. The CIFAR-10 dataset consists of $60000$ $32\times32$ colour images in 10 classes, with $6000$ images per class. To show the linear correlation of the values, we calculated the Pearson correlation coefficient (r) for each result presented. CIFAR-10, even though not directly related to O-RAN and the optimisation of a 5G network, was chosen due to the large availability of existing models from the community. The results for both setups were quite similar, and Fig.~\ref{fig:init_investigation} summarises the results for setup no.1.

Fig.~\ref{subfig:acc_energy}  displays the highest accuracy achieved compared to the total energy consumed. As it is evident, there is no direct correlation between these metrics ($r = 0.34$) with models consuming less energy and achieving higher accuracy (e.g., ResNet achieved $0.30\%$ higher accuracy than GoogleNet consuming $4\times$ less energy). Thus, sharing optimisation strategies across different models is rather difficult; each model requires its own optimisation. Fig.~\ref{subfig:train_energy} illustrates the results for energy against training time, which have a strong correlation (r = 0.999). This shows that time can be a helpful energy consumption indicator when energy readings from MSRs are unobtainable. It also implies that shorter experiments can still provide valid results when profiling different power limits, thus our decision for the \SI{30}{\second} profile intervals (as described in Sec.~\ref{subsec:energyMeasurements}).

Lastly, we plotted the average GPU utilisation against the average power draw (Fig.~\ref{subfig:util_power}). The results are calculated based on $P_\mathrm{tr} = \nicefrac{E_\mathrm{tr}}{T_\mathrm{tr}}$. The results indicate that utilisation and power are highly correlated but only up to a certain point. Beyond a power draw of \raisebox{-0.6ex}{\~{}}\SI{300}{\watt}, any further increase did not translate to GPU utilisation. At this point, utilisation was almost $100\%$, so performance was pretty much at its maximum. Both ResNeXt and PNASNet consumed more power than the other models, with ResNext obtaining 1\% of extra utilisation, but PNASNet achieved no benefit. These findings demonstrate that increased power does not always result in better model performance, which motivated the investigation of power capping and the proposed profiler.

\subsection{Overhead Analysis}
\begin{figure}[t]
  \centering
  \includegraphics[width=1\columnwidth]{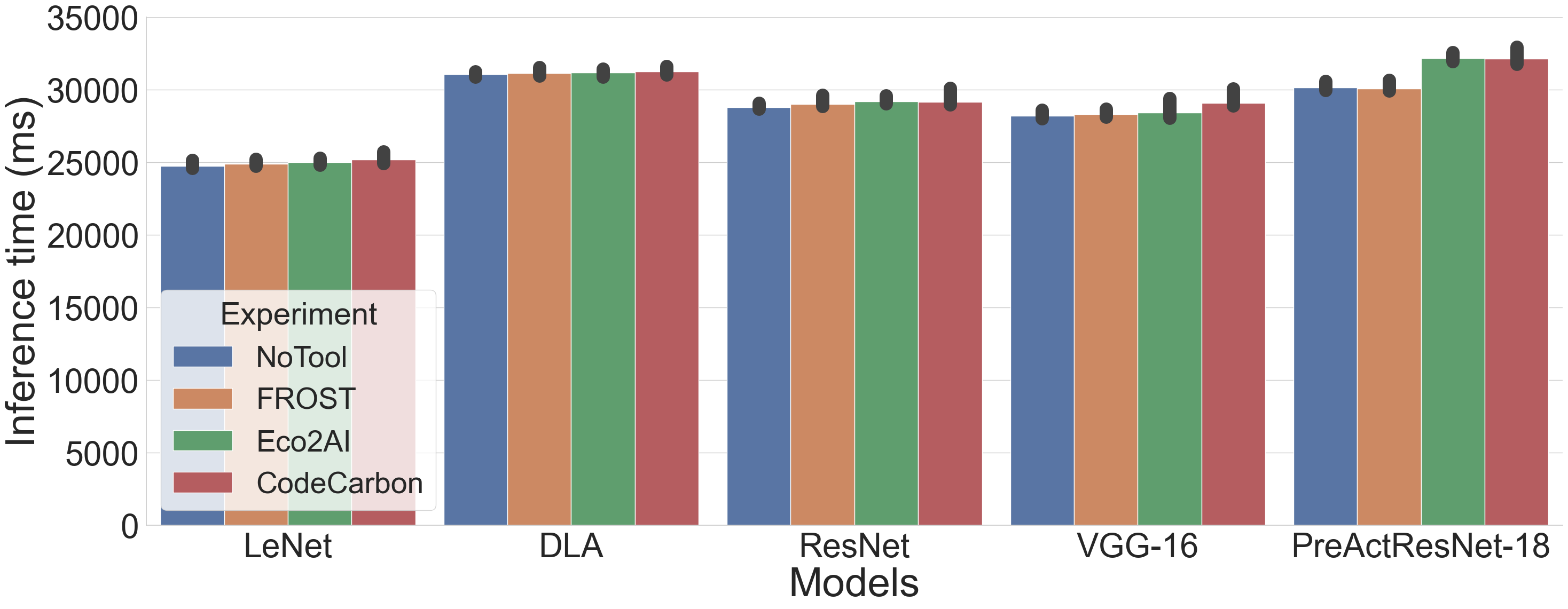}
  \caption{Overhead for FROST and similar tools in the literature.}
  \label{fig:overhead}
\end{figure}

The timely manner of the O-RAN operation dictates that any energy measuring technique should keep overhead to a minimum. Various tools are available for measuring energy, such as CodeCarbon~\cite{CodeCarbon} and Eco2AI~\cite{eco2ai}. In Fig.~\ref{fig:overhead},  we compare the time taken to infer across $50\mathrm{k}$ samples of the CIFAR-10 dataset using our implementation, CodeCarbon, Eco2AI, and a baseline experiment with no energy measurement. CodeCarbon and FROST use the same APIs, while Eco2AI uses Nvidia's NVML library for the GPU and introduces a generic CPU implementation. Both tools provide similar energy measurements with FROST, with the addition of various analytics related to carbon emissions.

As shown in (Fig.~\ref{fig:overhead}), our FROST implementation performs similarly to the baseline experiment, averaged across 100 experiments. However, CodeCarbon and Eco2AI, for specific models like VGG or PreActResNet, introduce a slight overhead. Our sampling rate was set at \SI{0.1}{\hertz}, while CodeCarbon and Eco2AI provide a sampling rate of \SI{1}{\hertz}. While our implementation collects more samples, the additional features provided by both frameworks can be attributed to the increased overhead they introduce. While this experiment was short and may not reflect the billions of samples parsed during inference in a real-world deployment, simplified implementations should be considered for large-scale O-RAN ML deployments to ensure smooth operation.

\subsection{Power Profiling Evaluation}
\begin{figure}[t]
  \centering
  \subfloat[MobileNet.]{\includegraphics[width=0.342\columnwidth]{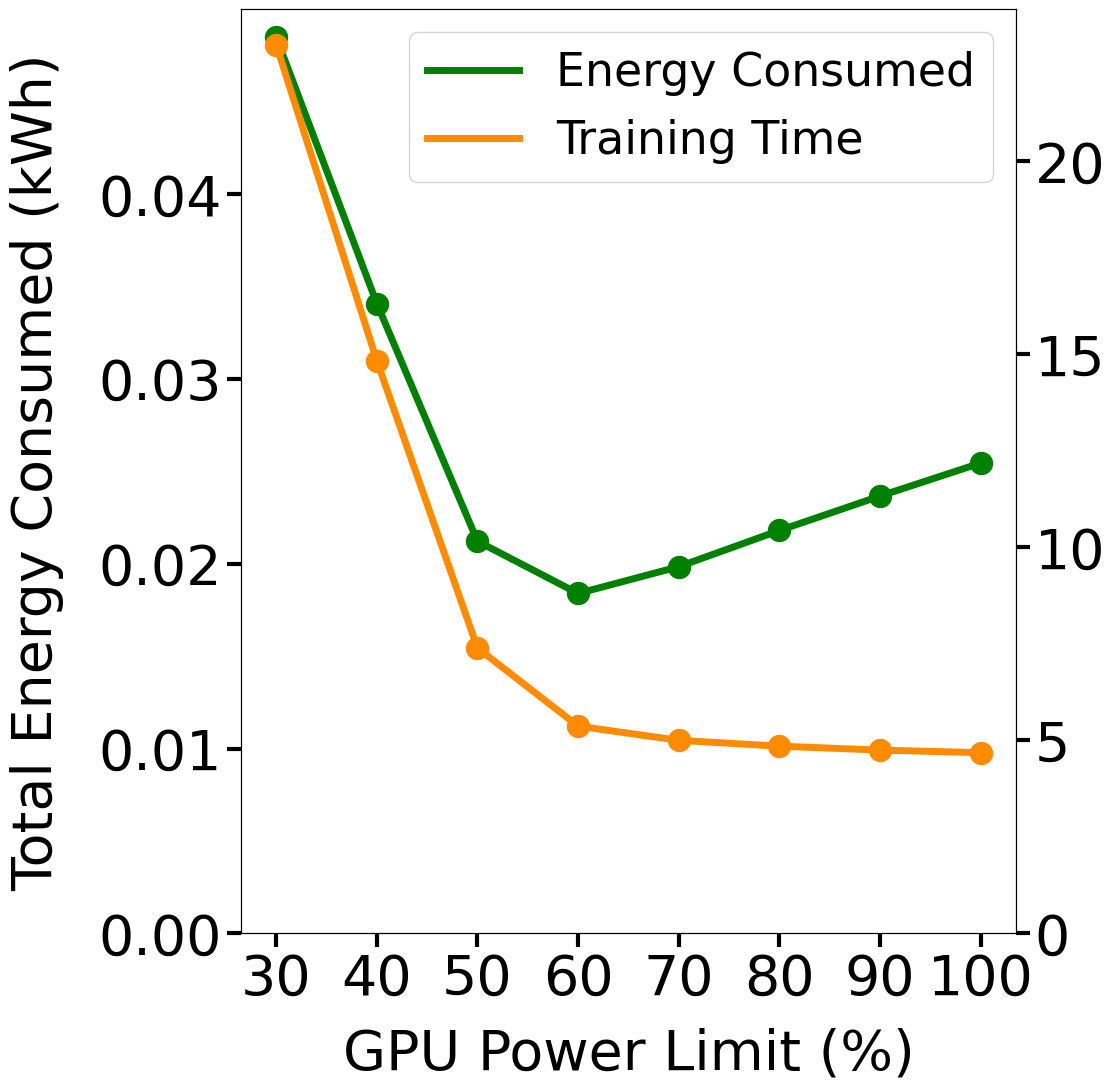}\label{subfig:mobilenet}}\hfill
  \subfloat[EfficientNet.]{\includegraphics[width=0.312\columnwidth]{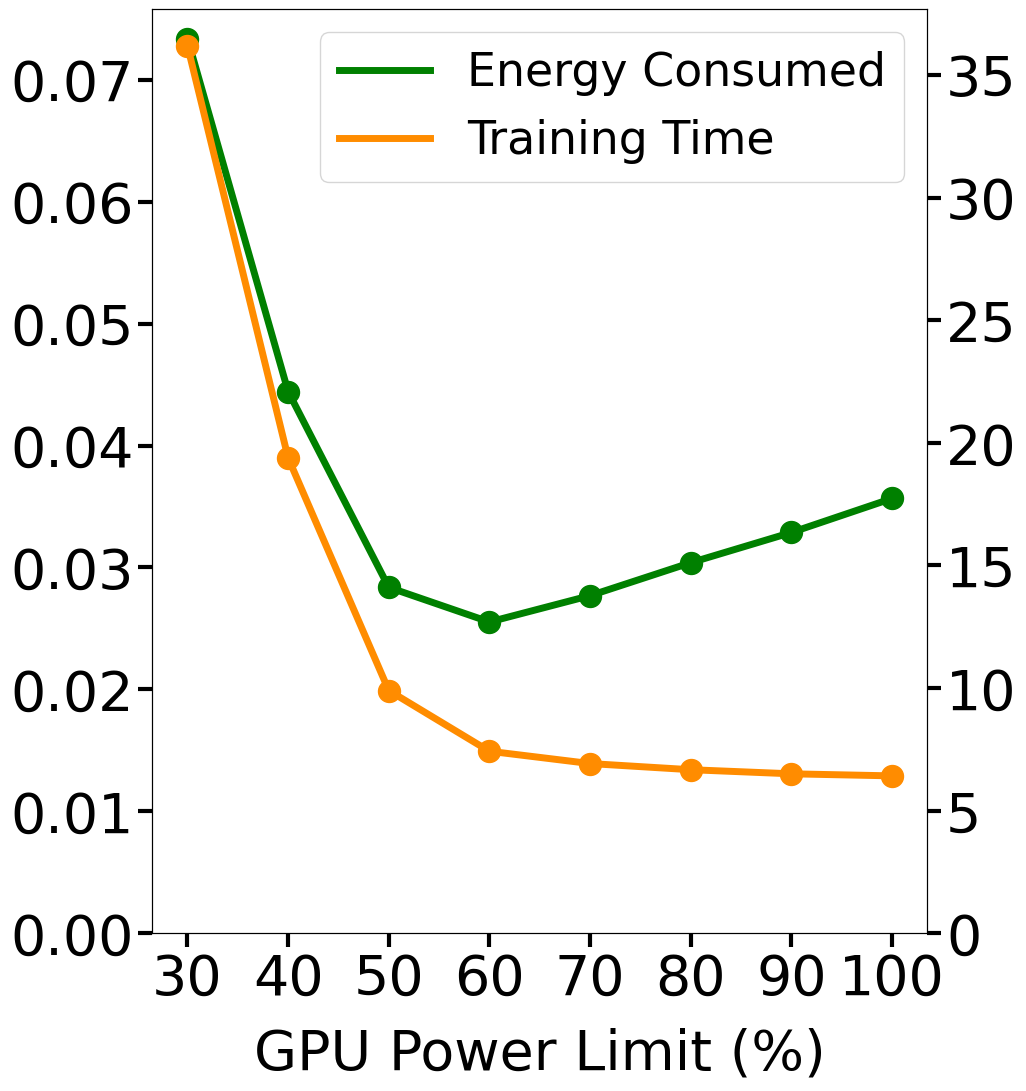}\label{subfig:efficientnet}}\hfill
  \subfloat[DPN.]{\includegraphics[width=0.336\columnwidth]{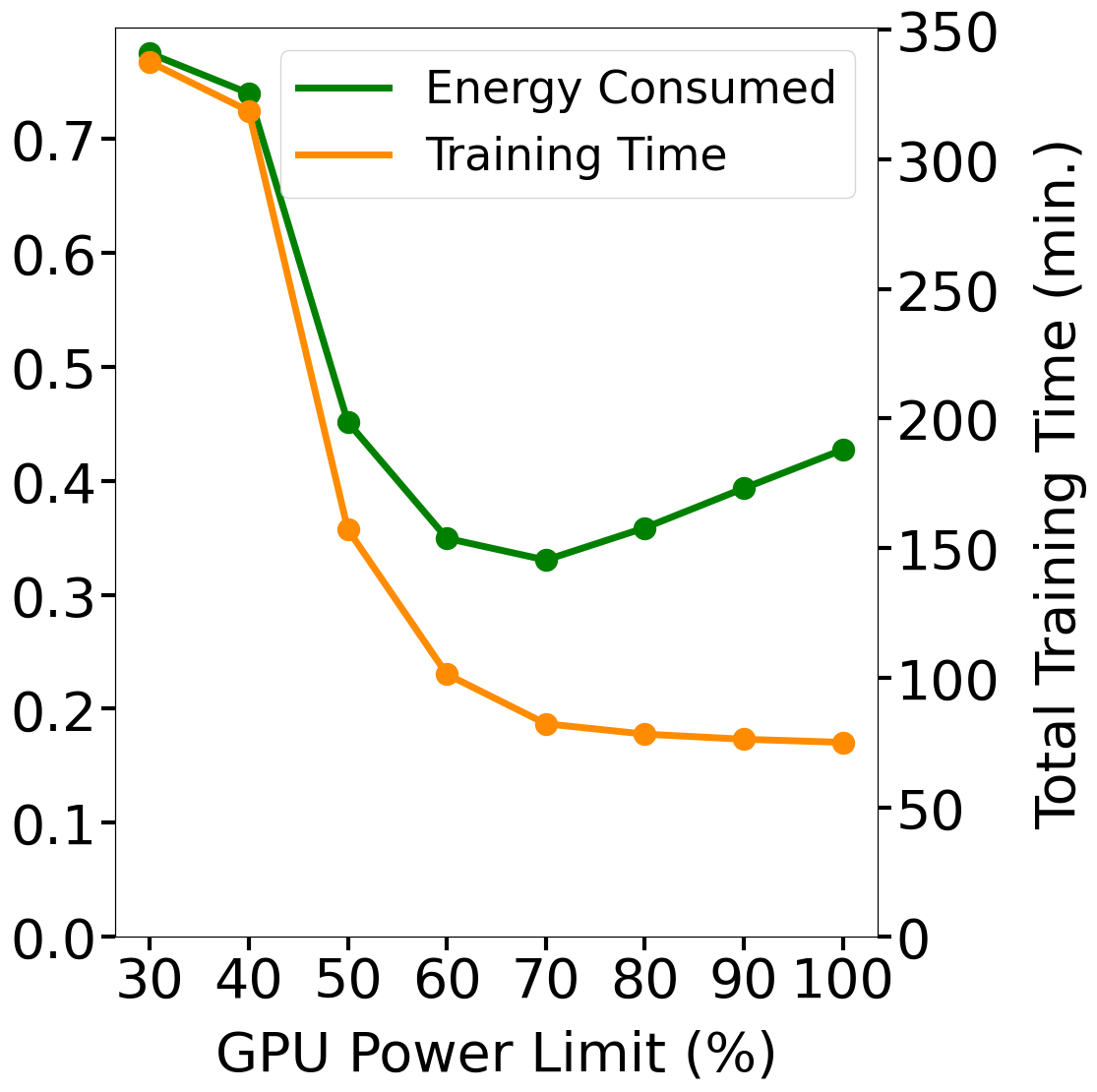}\label{subfig:dpn}}\hfill
  \caption{Example results of the power capping for three models.}
  \label{fig:power_capping}
\end{figure}

Our final investigation evaluated our power profiling optimisation strategy across all models. Due to limited space, we have included three example results from setup no.2 in Fig.~\ref{fig:power_capping}. We found that limiting power below the default profile of $100\%$ resulted in significant energy savings. Each model had an optimal power limit for energy consumption (e.g., MobileNet and DenseNet both had optimal limits of $60\%$, while EfficientNet had an optimal limit of $40\%$). Comparing results from both setups, we found that some models had different optimal limits (e.g., DPN had a minimal energy consumption limit of $60\%$ for setup no.1 and $70\%$ for setup no.2). LeNet was an outlier and showed no change in behaviour for both setups, likely due to the excessive power of the GPUs used. As discussed in Sec.~\ref{subsec:profiler}, there is a tradeoff between energy consumption and delay. This tradeoff is observed in Fig.~\ref{fig:power_capping}, but it was shown that energy reductions were more significant than delays for all models.

Introduce extreme capping (values less than $30\% - 40\%$)  can cause energy and time usage to increase sharply. Aggressively low limits can create instability in the GPU's circuitry, resulting in voltage fluctuations and improper functionality. Moreover, reducing the GPU clock frequency does not significantly affect runtime when power levels are higher, likely because the program is partially memory-bound. However, if the frequency becomes too low, the program becomes compute-bound, and the frequency becomes the bottleneck. This leads to a significant increase in execution time and energy consumption. Overall, a less aggressive power profile is always the best option for any hardware configuration.

\begin{figure}[t]
  \centering
  \includegraphics[width=1\columnwidth]{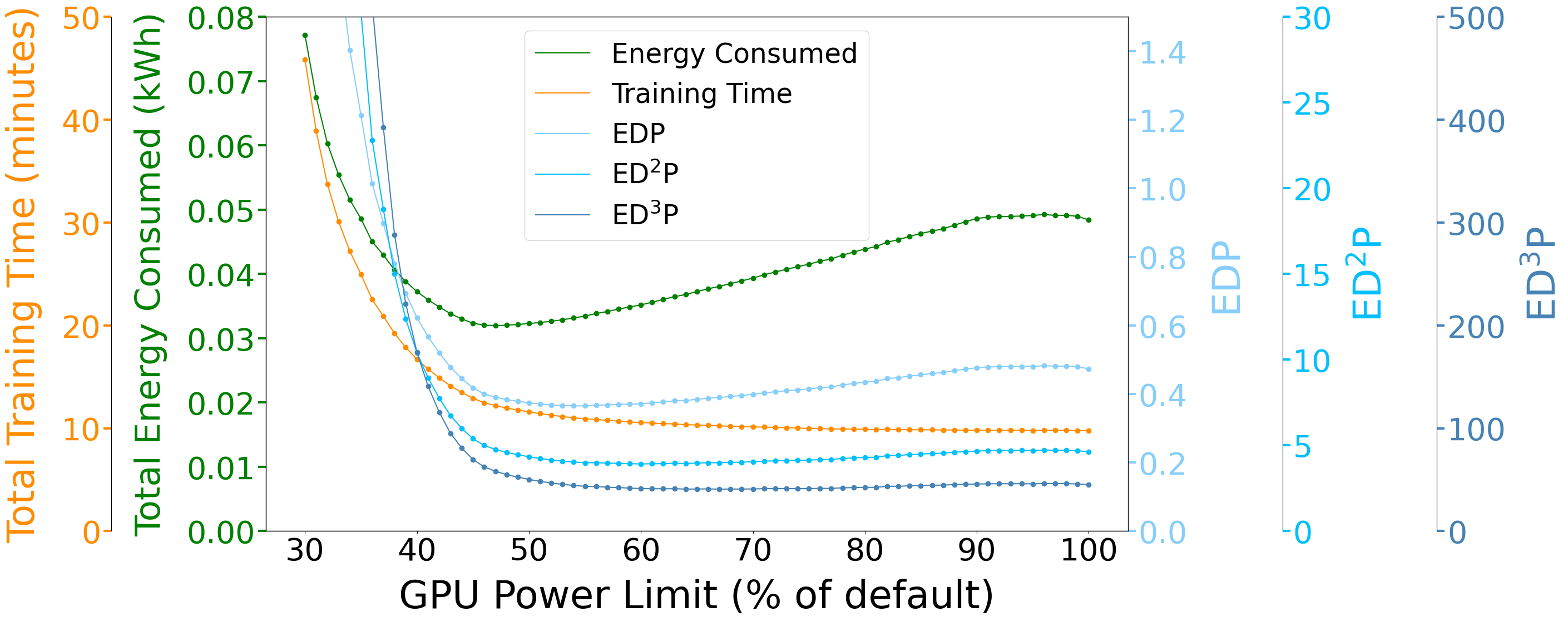}
  \caption{Fine-grained experiment for the ResNet model and different $\mathrm{ED^xP}$ optimisations.}
  \label{fig:edxp}
\end{figure}

\begin{figure}[t]
  \centering
  \includegraphics[width=1\columnwidth]{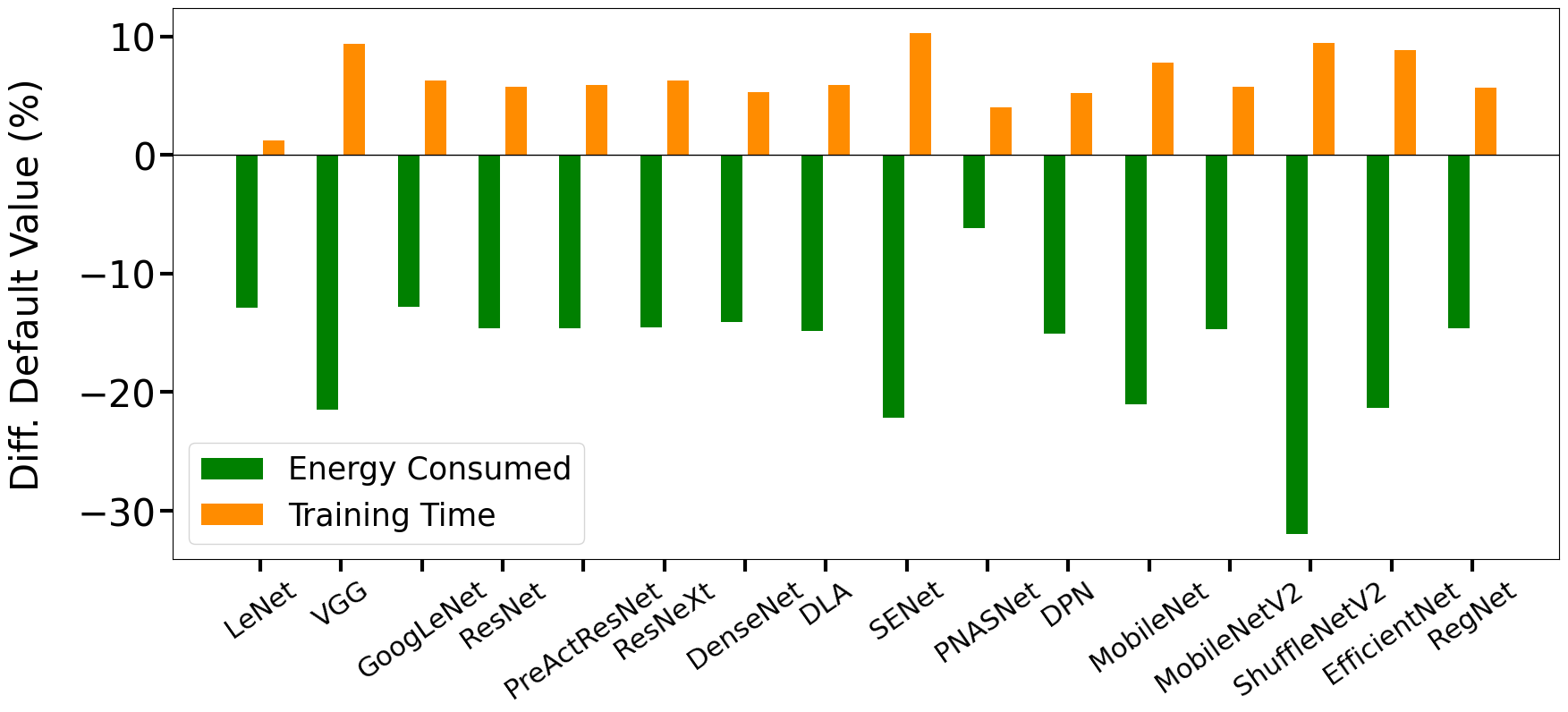}
  \caption{Overiew results showing the tradeoff between energy reductions and delay introduced.}
  \label{fig:powerlimit}
\end{figure}

Fig.~\ref{fig:edxp} presents an example of a fine-grained experiment. For this study, limiting power in increments of 1\%, we were able to observe the impact on energy and training time in greater detail. Our findings showed that power and frequency are directly proportional, but increasing frequency requires a corresponding increase in voltage to maintain stability. The equation $P = \nicefrac{1}{2}\  C V^2 f$ broadly applies to both CPUs and GPUs, showing that power and frequency are directly proportional. As voltage has a quadratic relationship to power, increasing frequency beyond a certain point leads to improved training times but significantly higher energy consumption. This is evident across all results (except LeNet) and motivated this study. 

We also evaluated different $\mathrm{ED^xP}$ decision-making criteria, as discussed in Sec.~\ref{subsec:profiler}. Fig.~\ref{fig:edxp} summarises these results. It was seen that the more weight attributed to delay, the higher the optimal power limit becomes. For $\mathrm{ED^3P}$, some of the optimal solutions were the maximum, implying that too much weight goes towards the delay, severely limiting the energy-saving benefits. $\mathrm{EDP}$ produced the greatest energy savings. Overall, for an O-RAN system, practitioners should adjust these parameters considering the unique QoS requirements of each application.

Our final figure (Fig.~\ref{fig:powerlimit} - setup no.1) provides an overview of the tradeoffs introduced by power capping. We found that $\mathrm{ED^2P}$ was the sweet spot between energy reductions and delay introduced (Fig.~\ref{fig:edxp}). On average, we observed a $26.4\%$ energy saving across all models for setup no.1, compared to $17.7\%$ for setup no.2. Similarly, training time increased by $6.9\%$ for setup no.1 and $5.5\%$ for setup no.2. Although our experimentation was limited to two setups, we believe that larger models may yield greater benefits. The more powerful RTX 3090 GPU of setup no.2 was utilised suboptimally with the evaluated models and hyperparameters used, and increased benefits can be seen if larger models are investigated (e.g., natural language models). Overall, power capping was shown to be effective in saving energy on both hardware setups and all models.

\section{Conclusions}\label{sec:conclusions}
In this study, we developed and tested strategies to optimise energy use for CNNs in the O-RAN context. Our research found no direct correlation between accuracy and energy consumption. However, there is a strong connection between energy use and training time, as well as between GPU utilisation and power draw. As a result, we proposed FROST, a power profiling strategy that utilises power limits and hardware optimisation. FROST has minimal overhead and is an efficient solution for O-RAN environments, resulting in significant energy savings across various models. We discussed the tradeoff between reducing energy and delaying operations, providing metrics tailored to specific uses and ensuring the necessary QoS for each application. Our approach proved effective across different hardware setups and models, resulting in average energy savings of $26.4\%$ and $17.7\%$ for the two distinct setups examined.

\section*{Acknowledgment}
This work was supported by Toshiba Europe Ltd. and Bristol Research and Innovation Laboratory (BRIL).

\bibliographystyle{IEEEtran}
\bibliography{bib}

\begin{thebibliography}{10}
\providecommand{\url}[1]{#1}
\csname url@samestyle\endcsname
\providecommand{\newblock}{\relax}
\providecommand{\bibinfo}[2]{#2}
\providecommand{\BIBentrySTDinterwordspacing}{\spaceskip=0pt\relax}
\providecommand{\BIBentryALTinterwordstretchfactor}{4}
\providecommand{\BIBentryALTinterwordspacing}{\spaceskip=\fontdimen2\font plus
\BIBentryALTinterwordstretchfactor\fontdimen3\font minus
  \fontdimen4\font\relax}
\providecommand{\BIBforeignlanguage}[2]{{%
\expandafter\ifx\csname l@#1\endcsname\relax
\typeout{** WARNING: IEEEtran.bst: No hyphenation pattern has been}%
\typeout{** loaded for the language `#1'. Using the pattern for}%
\typeout{** the default language instead.}%
\else
\language=\csname l@#1\endcsname
\fi
#2}}
\providecommand{\BIBdecl}{\relax}
\BIBdecl

\bibitem{website5G}
\BIBentryALTinterwordspacing
S.~Evans. {Global 5G Connections to Double by 2025}. [Online]. Available:
  \url{https://aibusiness.com/verticals/mwc-23-5g-connections-to-double-by-2025}
\BIBentrySTDinterwordspacing

\bibitem{5gRAN}
M.~A. Habibi, M.~Nasimi \emph{et~al.}, ``{A Comprehensive Survey of RAN
  Architectures Toward 5G Mobile Communication System},'' \emph{IEEE Access},
  vol.~7, pp. 70\,371--70\,421, 2019.

\bibitem{oranOverview}
A.~S. Abdalla, P.~S. Upadhyaya \emph{et~al.}, ``{Toward Next Generation Open
  Radio Access Networks: What O-RAN Can and Cannot Do!}'' \emph{IEEE Network},
  vol.~36, no.~6, pp. 206--213, 2022.

\bibitem{oranResiliency}
A.~Arnaz, J.~Lipman \emph{et~al.}, ``{Toward Integrating Intelligence and
  Programmability in Open Radio Access Networks: A Comprehensive Survey},''
  \emph{IEEE Access}, vol.~10, pp. 67\,747--67\,770, 2022.

\bibitem{5gMLapplications}
M.~E. Morocho-Cayamcela, H.~Lee, and W.~Lim, ``{Machine Learning for 5G/B5G
  Mobile and Wireless Communications: Potential, Limitations, and Future
  Directions},'' \emph{IEEE Access}, vol.~7, pp. 137\,184--137\,206, 2019.

\bibitem{gsmaIntelligence}
E.~Kolta, T.~Hatt \emph{et~al.}, ``{Going Green: Benchmarking the Energy
  Efficiency of Mobile},'' GSMA Intelligence, Tech. Rep., Jun. 2021.

\bibitem{5gRANEnergyReduction}
C.~B.~B. De~Souza, J.~J. Abularach~Arnez \emph{et~al.}, ``{Analysis of Power
  Consumption in 4G VoLTE and 5G VoNR Over IMS Network},'' in \emph{IEEE
  CAMAD}, Nov. 2022, pp. 59--64.

\bibitem{understandingORAN}
M.~Polese, L.~Bonati \emph{et~al.}, ``{Understanding O-RAN: Architecture,
  Interfaces, Algorithms, Security, and Research Challenges},'' \emph{arXiv},
  vol. abs/2202.01032, 2022.

\bibitem{oranEnergyEfficiencyCellSwitchOff}
G.~Vallero, D.~Renga \emph{et~al.}, ``{Greener RAN Operation Through Machine
  Learning},'' \emph{IEEE TNSM}, vol.~16, no.~3, pp. 896--908, 2019.

\bibitem{oranEnergyEfficiencyChannelSwitching}
M.~Hoffmann and P.~Kryszkiewicz, ``{Reinforcement Learning for Energy-Efficient
  5G Massive MIMO: Intelligent Antenna Switching},'' \emph{IEEE Access},
  vol.~9, pp. 130\,329--130\,339, 2021.

\bibitem{energyPolicyAAAI}
E.~Strubell, A.~Ganesh, and A.~McCallum, ``{Energy and Policy Considerations
  for Modern Deep Learning Research},'' in \emph{Proc. of AAAI}, vol.~34,
  no.~09, Apr. 2020, pp. 13\,693--13\,696.

\bibitem{greenAI}
R.~Verdecchia, J.~Sallou, and L.~Cruz, ``\BIBforeignlanguage{English}{{A
  Systematic Review of Green AI}},'' \emph{\BIBforeignlanguage{English}{WIREs
  Data Mining and Knowledge Discovery}}, 2023.

\bibitem{TYang2017-est}
T.-J. Yang, Y.-H. Chen \emph{et~al.}, ``{A Method to Estimate the Energy
  Consumption of Deep Neural Networks},'' in \emph{Proc. of IEEE ACSSC}, 2017,
  pp. 1916--1920.

\bibitem{inaccurateMethods}
T.-J. Yang, Y.-H. Chen, and V.~Sze, ``{Designing Energy-Efficient Convolutional
  Neural Networks Using Energy-Aware Pruning},'' in \emph{Proc. of IEEE CVPR},
  2017, pp. 6071--6079.

\bibitem{oranMLFlow}
{O-RAN Working Group 2}, ``{O-RAN AI/ML Workflow Description and
  Requirements},'' O-RAN Alliance, Tech. Rep., Jul. 2021.

\bibitem{gpucapping}
A.~Krzywaniak, P.~Czarnul, and J.~Proficz, ``{GPU Power Capping for
  Energy-Performance Trade-Offs in Training of Deep Convolutional Neural
  Networks for Image Recognition},'' in \emph{Proc. of ICCS}.\hskip 1em plus
  0.5em minus 0.4em\relax Springer International Publishing, Jul. 2022, pp.
  667--681.

\bibitem{oranWhitePaper}
C.~Li and A.~Akman, ``{O-RAN Use Cases and Deployment Scenarios: Towards Open
  and Smart RAN},'' \emph{{O-RAN Alliance}}, 2020.

\bibitem{DVFS}
Z.~Tang, Y.~Wang \emph{et~al.}, ``{The Impact of GPU DVFS on the Energy and
  Performance of Deep Learning: An Empirical Study},'' in \emph{Proc. of ACM
  e-Energy}.\hskip 1em plus 0.5em minus 0.4em\relax ACM, 2019, p. 315–325.

\bibitem{amdWork}
F.~Mendes, P.~Tomás, and N.~Roma, ``{Decoupling GPU Voltage-Frequency Scaling
  for Deep-Learning Applications},'' \emph{J. Parallel Distrib. Comput.}, vol.
  165, pp. 32--51, 2022.

\bibitem{physicalMeter}
G.~Conti, D.~Jimenez \emph{et~al.}, ``{A Multi-Port Hardware Energy Meter
  System for Data Centers and Server Farms Monitoring},'' \emph{Sensors},
  vol.~23, no.~1, 2023.

\bibitem{Nvidia2016}
\BIBentryALTinterwordspacing
{NVIDIA Corporation}, ``{nvidia-smi.txt},'' 7 2016. [Online]. Available:
  \url{https://developer.download.nvidia.com/compute/DCGM/docs/nvidia-smi-367.38.pdf}
\BIBentrySTDinterwordspacing

\bibitem{HDavid2010}
H.~David, E.~Gorbatov \emph{et~al.}, ``{RAPL: Memory power estimation and
  capping},'' in \emph{Proc. of ACM/IEEE ISLPED}, 2010, pp. 189--194.

\bibitem{dramPowerConsumption}
T.~Vogelsang, ``{Understanding the Energy Consumption of Dynamic Random Access
  Memories},'' in \emph{Proc. of IEEE/ACM MICRO}, 2010, pp. 363--374.

\bibitem{edp}
R.~Gonzalez and M.~Horowitz, ``{Energy Dissipation in General Purpose
  Microprocessors},'' \emph{IEEE JSSC}, vol.~31, no.~9, pp. 1277--1284, 1996.

\bibitem{CodeCarbon}
\BIBentryALTinterwordspacing
``{CodeCarbon}.'' [Online]. Available:
  \url{https://github.com/mlco2/codecarbon}
\BIBentrySTDinterwordspacing

\bibitem{eco2ai}
S.~Budennyy, V.~Lazarev \emph{et~al.}, ``{Eco2AI: Carbon Emissions Tracking of
  Machine Learning Models as the First Step Towards Sustainable AI},'' in
  \emph{Doklady Mathematics}.\hskip 1em plus 0.5em minus 0.4em\relax Springer,
  2023, pp. 1--11.

\end{thebibliography}

\end{document}